# Towards artificially intelligent recycling: Improving image processing for waste classification

Youpeng Yu
*Electronic and Electrical Engineering*
*University College London*
London, England
uceeyyy@ucl.ac.uk

Ryan Grammenos
*Electronic and Electrical Engineering*
*University College London*
London, England
r.grammenos@ucl.ac.uk

*Abstract*— **The ever-increasing amount of global refuse is overwhelming the waste and recycling management industries. The need for smart systems for environmental monitoring and the enhancement of recycling processes is thus greater than ever. Amongst these efforts lies IBM's Wastenet project which aims to improve recycling by using artificial intelligence for waste classification. The work reported in this paper builds on this project through the use of transfer learning and data augmentation techniques to ameliorate classification accuracy. Starting with a convolutional neural network (CNN), a systematic approach is followed for selecting appropriate splitting ratios and for tuning multiple training parameters including learning rate schedulers, layers freezing, batch sizes and loss functions, in the context of the given scenario which requires classification of waste into different recycling types. Results are compared and contrasted using 10-fold cross validation and demonstrate that the model developed achieves a 91.21% test accuracy. Subsequently, a range of data augmentation techniques are then incorporated into this work including flipping, rotation, shearing, zooming, and brightness control. Results show that these augmentation techniques further improve the test accuracy of the final model to 95.40%. Unlike other work reported in the field, this paper provides full details regarding the training of the model. Furthermore, the code for this work has been made open-source and we have demonstrated that the model can perform successful real-time classification of recycling waste items using a standard computer webcam.**

## I. INTRODUCTION

Waste management has become a global challenge due to the increasing amount of world waste generation. The world bank group estimated that the world waste generation would experience a 70 percent increase between 2016 and 2050, from 2.01 billion to 3.40 billion tonnes [1]. To deal with this large amount of waste generation, image classification technologies can help improve the recycling process by introducing automated waste sorting processes. Such an example is the IBM Wastenet project [2] which looked at designing an Artificial Intelligence (AI) recycling bin, which helps people place rubbish into the correct bin.

The United Kingdom (UK) Government defines waste as materials discarded by the producers or holders. More specifically, discarding includes activities such as throwing away and recycling [3]. In general, waste can be classified into five types amongst which “packaging waste and recyclables” is the focus of IBM Wastenet. The Wastenet project aims at improving the UK’s low households’ recycling rate (45.5% in 2017) [2]. Recyclables can be defined as waste materials which are reprocessed into products through recycling and recovery operations, and common recyclables are paper, cardboard, plastic, metal, wood, and glass. Waste must be classified before sending it for recycling since the recycling process can vary depending on the recycling waste types [4].

The authors in [5] categorize automated sorting techniques for recycling waste into direct sorting and indirect sorting methods. Direct sorting methods can recognize and separate different recycling waste types using material properties only. On the other hand, indirect sorting methods can only identify different recycling waste types but need the help of robotic arms to separate them. The IBM Wastenet project performs recycling waste sorting using indirect sorting method – “optic based sorting”. Optic based sorting method utilizes camera-based sensors to take waste images, and identify recycling waste types through visual cues, such as colour, shape, and texture. The development of machine learning enables image-based recycling waste classification at the edge with various advantages: better privacy and safety, lower costs, and faster predictions.

The current approach to image-based waste classification is implementing transfer learning techniques on different well-known Convolutional Neural Network (CNN) architectures. To be specific, these well-known CNN architectures obtained great classification results in ImageNet Challenges (ILSVRC), where ILSVRC is a large-scale object recognition competition. ILSVRC models are trained on the ImageNet dataset, which contains over 14 million images across 21 thousand classes [6]. CNN architectures such as GoogleNet, VGG, and AlexNet are excellent choices of pre-trained models since they have successfully captured general features of various objects. Next, these CNN architectures with pre-trained weights are retrained on the waste image datasets. The most commonly used waste image dataset is TrashNet, which was created in 2017 and consists of 2527 waste images across six classes [7].

Significant research has been carried out in waste classification by applying transfer learning techniques to CNN models, and these research projects have achieved great classification results with over 95% test accuracy. However, limited research has been done in developing a systematic waste classification model training approach. More specifically, some research papers do not provide full model training details, and most of these papers do not justify their model training decisions.

The work reported in this paper builds on the IBM Wastenet project from the image processing aspect to improve the classification accuracy. The code developed during this work has been made open-source and available on the GitHub project repository [8]. A demonstration video of real-time classification has been recorded and available on YouTube [9]. The following contributions are presented within this paper: development of a systematic waste classification model training approach; provision of full model training details and justification for the design decisions made; selection of the most appropriate dataset splitting parameters through a comprehensive statistical analysis; selection of appropriate freezing layers and data augmentation techniques using 10-fold Cross Validation (CV).

This paper is organized as follows: Section II reviews related work in the field; Section III introduces the system architecture with the methodology and results presented in Section IV and Section V, respectively; Section VI concludes this paper and makes recommendations for future work.

## II. Related work

The use of machine learning algorithms extends to a myriad of real-world applications including gesture recognition [10] and increasingly in Internet of Things scenarios [11]. While a range of data analytics techniques have been considered to optimize waste management procedures [12], the work in this paper focuses specifically on the use of machine learning techniques to improve waste classification accuracy.

In 2017, the authors in [13] collected the first public recycling waste image dataset, called TrashNet, which contains 2527 images across six recycling types, including paper, plastic, glass, cardboard, metal, and general trash. An eleven-layer network that is similar to the AlexNet model with different numbers of layers and nodes was developed by the authors to perform waste image classification tasks, and achieved 22% classification accuracy. The authors later improved the model accuracy to around 75% by using the Kaiming initialization method.

In 2018, the authors in [14] compared the classification performance of different machine learning algorithms on the TrashNet dataset. These algorithms include CNN, Support Vector Machines (SVM), Random Forest (RF), and K-Nearest Neighbour (KNN). Among these algorithms, the CNN trained model achieves the highest classification accuracy, 89.81%. In the same year, the authors in [15] implemented transfer learning techniques on different well-known CNN architectures, including ResNet50, MobileNet, InceptionResNetV2, DenseNet121, and Xception. The best-performed model, DenseNet121, is retrained using the TrashNet dataset with pre-trained weights and achieved 95% test accuracy. Moreover, the authors in [16] used the GoogleNet model as the fine-tuned model and obtained the highest classification accuracy, 97.86%, on the TrashNet dataset by far.

In the last two years, the authors in [17]–[19] did not achieve higher classification accuracy, but made their contributions on testing different CNN models, training parameters, and data augmentation techniques. Overall, recent waste classification projects have achieved classification results with over 95% accuracy, by applying transfer learning and data augmentation techniques. However, the authors have not provided details of the model training undertaken. This paper follows a systematic approach to split datasets, tune training parameters, and choose data augmentation techniques while providing details of the model's training.

## III. System architecture

The best-performed model in [16] is reproduced to be the benchmark model for two reasons. Firstly, this model achieves the highest classification accuracy (97.86%). Secondly, data augmentation techniques are not applied to this model, which makes it a good choice to test various data augmentation techniques. However, this model's paper does not provide model training details other than the pretrained model's name (GoogleNet), final classifier used (Softmax and Support Vector Machine), and the dataset used (TrashNet), which makes it difficult to reproduce. To solve this problem, various data splitting methods and training parameters are tested, and the ones which give the highest classification accuracy are selected.

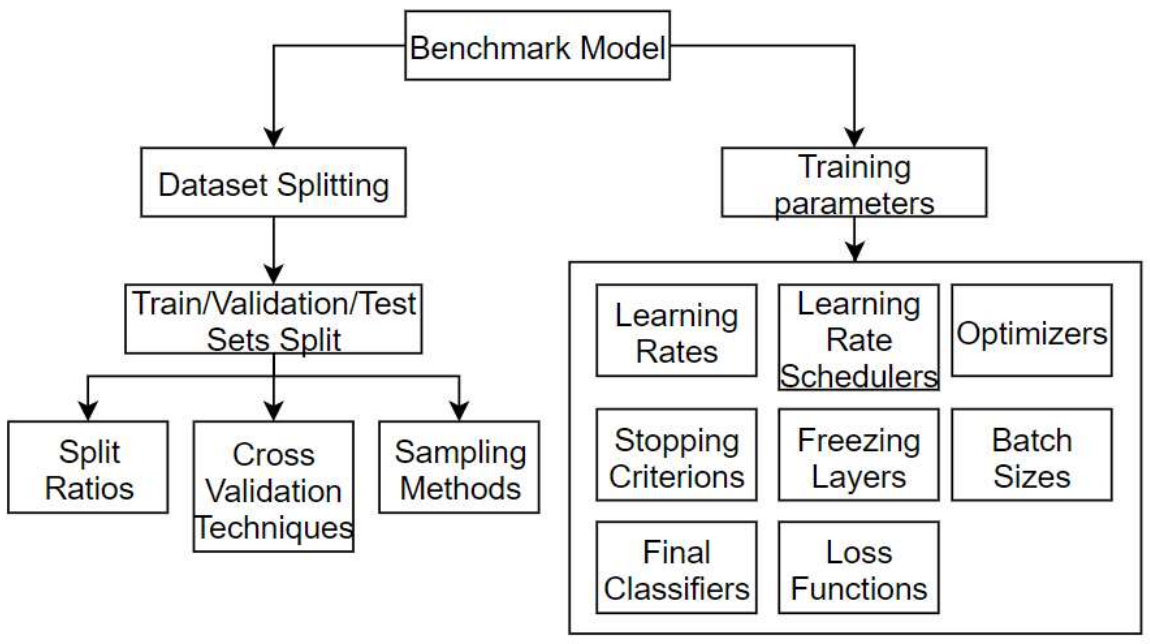

Fig. 1. Flowchart of developing a benchmark model

Figure 1 summarises the training procedure of developing the benchmark model for this work. Firstly, the TrashNet dataset is split into train, validation, and test sets using selected split ratios, cross validation techniques, and sampling strategies. Split ratios manage the number of waste images in each set, and sampling strategies decide the chance of each image being selected. Cross validation techniques control different partitions of a dataset, whereas a model's performance can be reflected more accurately by averaging validation results over multiple partitions.

Next, training parameters are tuned to achieve the highest classification accuracy, and to produce the best-performed benchmark model. Final classifiers and loss functions together calculate losses of a model's prediction on training set images. Batch sizes enable a model to predict and evaluate losses for multiple images in one round. Optimisers, learning rates, and learning rates schedulers together reduce losses. Stopping criterions set an acceptable range of losses for stopping the training process. Additionally, freezing layers help retrain the pre-trained model to fit to the TrashNet dataset, both retaining basic image information (such as shapes, colours, and lines) and learning to recognize different recyclables.

Lastly, various data augmentation techniques are tested on the reproduced benchmark model, and the set of techniques yielding the highest classification accuracy is applied to train the final model.

## IV. Methodology

Reproducing the model of paper [16] as the benchmark model was a challenge given the lack of training details. The overall approach to this research problem is to summarise and test parameter values used in other recycling waste classification projects, selecting the ones that provide the highest classification accuracy. The desired outcome of this work is to achieve a classification accuracy of 95% or higher, similar to other work reported in the field.

In comparison, using parameter values of other successful waste classification projects should give better prediction performance than randomly selected parameter values. However, some parameters are not discussed in these project papers, and most parameter values are selected without justification. To solve this problem, parameter values of multiple projects are summarised and tested to select appropriate parameter values. In this case, parameter values of seven projects [13]–[19], discussed in Section II, are summarised and tested. Next, specific approaches of splitting datasets, tuning training parameters, and applying data augmentation techniques are discussed.

Starting with datasets splitting, the test set of the TrashNet dataset should be split first and kept aside for the final model performance evaluation. Therefore, it can only be split using the Hold-out CV for confidentiality consideration. Then, the split ratio and sampling strategy can be determined by testing values used in reviewed projects. These adjusted dataset split ratios for train, validation, and test sets are 80/10/10, 70/15/15, 60/20/20, and 50/25/25 in percentage, and sampling strategies are Simple Random Sampling and Stratified Random Sampling. Next, K-fold CV is used to ensure a low model bias when splitting the remaining dataset into train and validation sets, and the K value also determines the split ratio of two sets. In this case, K=5 and K=10, two most used K values in reviewed projects are tested.

Dataset splitting parameters including split ratios, sampling strategies, and CV techniques are selected based on the mean comparison test results. To elaborate on that, a dataset can have many different splitting patterns given the same split ratio, and different splitting patterns trained model will have different classification accuracy. The dataset split ratio decision can be biased if only run each split ratio experiment (split a dataset using the given split ratio and train a model using this split dataset) once and compare their classification accuracies. Instead, running each split ratio experiment multiple times and performing mean comparison tests on the experimental results can help reduce the effects of randomness.

Furthermore, the number of experiment times for each split ratio is described as the sample size. Sample sizes are determined if the estimated standard error of the mean of four split ratios are reduced to the same level, where the standard error of the mean represents how far the sample mean deviates from the population mean and can be reduced by repeating experiments. After that, normality and homogeneity of variances tests are carried out to choose between parametric and non-parametric sample mean comparison tests. Lastly, mean comparison test results are evaluated and the ones that give highest estimated population mean accuracy are selected.

Subsequently, various training parameters are summarised from the reviewed projects and tested, and the ones that give higher average accuracy are selected. More specifically, learning rate schedulers options are constant learning rate, decay learning rate, and cyclic learning rate; optimizers are stochastic gradient descent (SGD), Adaptive moment estimation (Adam), and Adadelta; final classifiers are Softmax, Linear SVM, and Non-linear SVM; loss functions are cross entropy loss and hinge loss; batch sizes are 8, 16, and 32. Patience epochs and learning rates are identified using the highest accuracy epoch gap and learning rate finder, respectively. Additionally, freezing layers are determined by unfreezing the model from the top classifiers to the bottom input layers.

Finally, various data augmentation techniques are summarised from reviewed projects and tested, and the set that gives the highest average validation accuracy is applied to train the final model. These data augmentation techniques are flipping (horizontal and vertical), rotation (15, 40, 90, and 180 degrees), shear (1, 10, 30, 60, and 89 degrees), zoom (25, 50, and 100 percent), and brightness control (10, 25, and 50 percent). Confusion matrices are plotted to analyse the effects of different data augmentation techniques on different types of recyclables.

## V. Results

The research problem, which involves reproducing the model of paper [16] with data augmentation techniques applied, is solved by conducting experiments on dataset splitting, training parameters tuning, and data augmentation techniques choosing. In this section, these experiment results are presented, analysed, and discussed in subsections *A* (*Dataset split*), *B (Training parameters)*, and *C (Data augmentation)*.

TABLE I. The final model settings

| | CV techniques | split ratios | sampling strategies |
|---|---|---|---|
| **Training set** | 10-fold CV | 81% | Simple Random Sampling |
| **Validation set** | | 9% | |
| **Test set** | Hold-out CV | 10% | |
| | | | |
| | **Training parameters** | | |
| **Learning rate scheduler** | Constant learning rate | | |
| **Optimiser** | Adam optimiser | | |
| **Stopping criteria (patience epochs)** | 100 epochs | | |
| **Layers freezing** | Freeze base convolutional layers | | |
| **Loss function** | Cross Entropy loss | | |
| **Final classifier** | Softmax classifer | | |
| **Batch size** | 16 | | |
| **Learning rate** | 2.00E-05 | | |
| | | | |
| | **Data augmentation techniques** | | |
| **Flipping** | Horizontal Flipping+Vertical Flipping | | |
| **Rotation** | 180 degrees | | |
| **Shear** | 89 degrees | | |
| **Zoom** | 100% | | |

Table I presents the dataset split options, training parameters, and data augmentation techniques that are used to train the final model of this work. To elaborate on the dataset split settings, 10% of the entire dataset is separated into the test set using simple random sampling strategy. Then, 10-fold CV is applied and split the rest dataset (90% of the entire dataset) into training and validation sets. More specifically, nine folds are combined and used as the training set each time, and the remained fold is used as the validation set. Lastly, simple random sampling strategy is applied to both Hold-out CV and 10-fold CV.

Next, all base convolutional layers of the GoogleNet model are frozen. Accordingly, a small learning and a small batch size are selected to protect the low-level features extracted from the pre-trained weights. Classic CNN training parameters, such as constant learning rate scheduler, Adam optimizer, Cross-Entropy, and Softmax final classification layer, are used to train the final model. Lastly, four data augmentation techniques, except the brightness control, are applied to mitigate the effects of the overfitting problem, and to achieve higher classification accuracy.

## *A. Dataset Split*

The TrashNet dataset is split into train, validation, and test set to train CNN model, and the methods of choosing splitting parameters for each set are similar. The selection of the test set split ratio is presented in this subsection for illustration, which involves identifying appropriate sample sizes, performing data distribution analysis and mean comparison tests.

### *1) Sample size*

Standard errors of the sample mean of four test split ratios (10%, 15%, 20%, and 25%) are recorded with respect of sample sizes, and four split ratio curves are plotted on the same graph to identify the appropriate sample size for each split ratio.

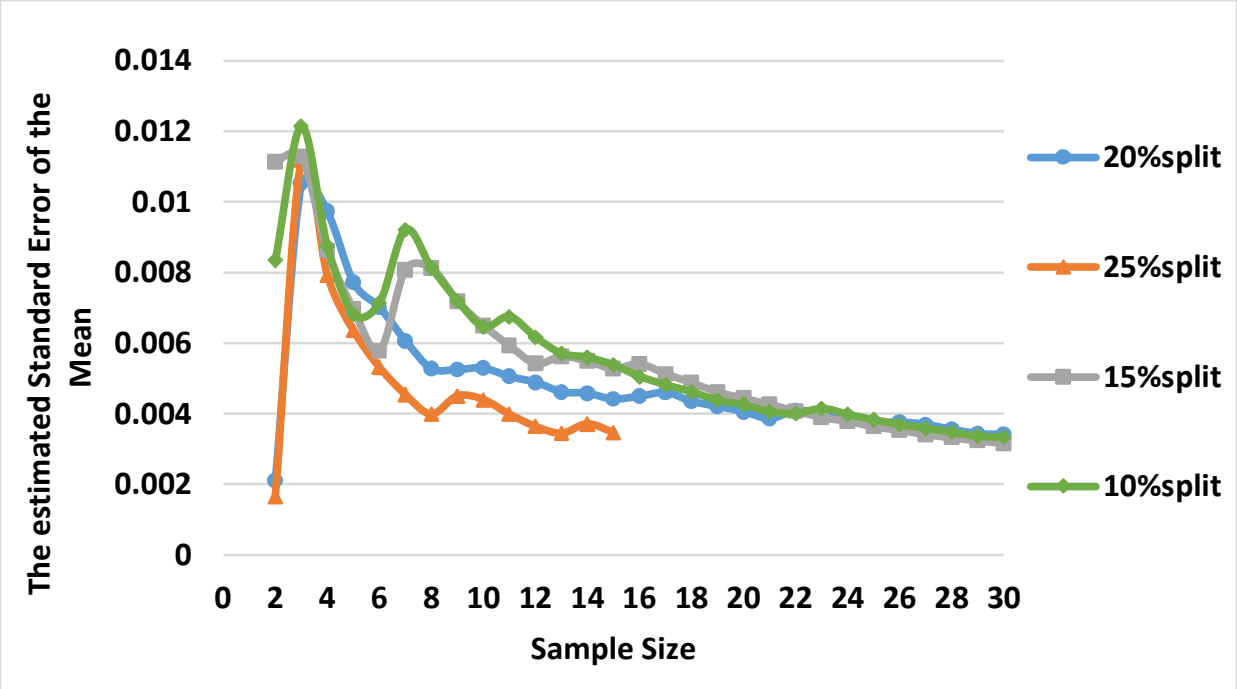

Fig. 2. The estimated standard error of the mean against sample sizes

From Figure 2, the estimated standard error of the mean of all four split ratios are reduced to the same level, slight above 0.003. These low standard errors indicate that the sample mean of the current sample sizes has well approximated the population mean. In other words, the current sample sizes are selected large enough. In this case, appropriate sample size selections are 15 for the 25% split and 30 for the other split ratios.

$$\sigma_{\bar{x}} \approx \frac{s}{\sqrt{n}} \quad (1)$$

As can be seen from Equation 1, the estimated standard error of the mean is inversely proportional to the sample size, where $\sigma_{\bar{x}}$ is the standard error of the mean, $s$ is the sample standard deviation, and $n$ is the sample size. The relationship identified in Equation 1 can be viewed in Figure 2, where the estimated standard errors of all four splits show decreasing trends with increasing sample sizes.

$$s = \sqrt{\frac{\sum(x_i - \bar{x})^2}{n-1}} \quad (2)$$

In Equation 2, $s$ is the sample standard deviation, $x_i - \bar{x}$ is the difference between sample values and the population mean, and $n$ is sample size. From Figure 2, the estimated standard errors of the mean are not defined when sample size=1 because of Bessel's correction [20]. More specifically, $n-1$ is used instead of $n$ to correct the bias of estimating the population standard deviation. Next, 20% and 25% split curves start (sample size=2) with low estimated standard errors, and increase rapidly to the maximum at the next point (sample size=3). From Equation 1, the estimated standard error should be large when the sample size is small since they are inversely proportional. The authors in [21] pointed out that the standard error from a small sample size is likely to underestimate the standard error of the mean. In other words, the first few randomly collected samples can be very close to the mean and therefore a small standard deviation value, but can be corrected as more samples are collected.

Then, steep decreases can be observed in the next few increases of sample size in all splits. To be specific, 64.3% and 50% standard error decreases can be observed in 20% split and 25% split respectively when the sample size is increased from 3 to 8. On the other hand, steep standard error decreases can only be observed before sample size of 5 in 10% split and sample size of 6 in 15% split. This suggests that large test set split ratio needs fewer sample sizes to have low estimated standard errors, and can be explained by the total number of samples. To be specific, the 25% split with sample size=15 can collect 9476 total sample images given the TrashNet has 2527 images, and the 15% split need to have sample size=25 to collect the same amount of total sample images. Therefore, a smaller sample size is selected for the 25% split compared with the other split ratios.

After that, the estimated standard errors of all four splits decrease slowly and become stable. To be specific, the increase of the sample size cannot further reduce the estimated standard error significantly. In other words, the standard error gradually decreases as more samples are collected, and it converges to zero when the sample mean approximates to the population mean.

The significance of Figure 2 is to identify the relationship between the standard error of the sample mean and sample size, and to help choose an appropriate sample size for each split ratio. More specifically, an appropriate sample size selection should not be too small that the standard error is large, also not too large that the reduction of standard error becomes too small. These experimental results suggest that the estimated standard error of the sample mean is a good indicator of selecting the sample size.

### *2) Statistical analysis*

Next, statistical analysis is performed on the collected samples to compare the sample mean of four split ratios. analysis of variance (ANOVA), a parametric test, is often used to compare means between multiple groups. The Shapiro-Wilk test and Levene's test are used to testing ANOVA's two assumptions, normal distribution and homogeneity of variances, respectively. One-tailed t-test, a pair-wise sample mean comparison test, is then performed to compare sample means of two split ratios. A commonly used confidence level, 0.95, is used throughout these tests.

- The p-value of Shapiro-Wilk test (confidence level of 0.95) is 0.53, 0.70, 0.44, and 0.99 in 10%, 15%, 20%, and 25% split, respectively.
- The p-value of Levene's test (confidence level of 0.95) is 0.27.
- The p-value of ANOVA test (confidence level of 0.95) is 0.0036.
- The p-value of pairwise one-tailed t-test (confidence level of 0.95) is 0.059 between 10% and 15%, 0.037 between 10% and 20%, and <0.001 between 10% and 25%.

The Shapiro-Wilk tests' results suggest that these four splits data are normally distributed since their p-values are much larger than 0.05. Next, the p-value of Levene's test is larger than 0.05, which proves that there is not a significant difference in variance between these four splits. With two assumption hold, the ANOVA test is valid. The ANOVA test's result suggests that at least one group's mean is different from others since the p-value is smaller than 0.05. Additionally, one-tailed t-tests are performed pairwise between the highest sample mean group 10% and other split groups to identify mean differences. There is not a significant difference between 10% and 15% split since their t-test p-value is larger than 0.05. On the contrary, 10% split has larger mean compared with 20% split and 25% split since their p-values are smaller than 0.05. As a result, the difference between sample mean accuracy of 10% split and 15% split is not significant, and both split ratios are appropriate choices.

Statistical analysis is useful to determine which dataset split ratio gives the highest classification mean accuracy. In comparison, the sample mean accuracy can be biased and misleading since the sample size is not large enough. The ANOVA test takes the sample size into account and compares the sample mean, which conduct a more accurate analysis. In this project, the split ratio of the test set is selected by comparing mean classification accuracy using statistical tests, which makes the experiment result has statistical significance. This approach of selecting the test set split ratio can also be applied to other recycling waste classification projects to split the dataset and improve classification accuracy.

#### *3) The other dataset splitting parameters*

The other dataset splitting parameters including the sampling strategy used in the test set and splitting parameters used in train and validation sets (the K-value of K-fold CV, the sampling strategy, and the split ratio) can also be determined using the approach discussed in subsection *A1* and *A2*. Statistical analysis proof of the other dataset splitting decisions are not discussed here since they follow the same approach. To summarise, determine the sample size of all candidate parameter values using the estimated standard error of the sample mean, and perform statistical analysis to choose between dataset splitting parameters.

### *B. Training parameters*

Various training parameters are tuned based on the 10-fold CV average accuracy. These training parameters are learning rate schedulers, optimisers, patience epochs, layers freezing, loss function, final classifiers, batch sizes, and learning rates.

#### *1) Effects of tuning different training parameters*

Various training parameters are tuned to help the final model achieve higher classification accuracy. The average validation accuracy is plotted as a curve to identify the effects of different training parameters.

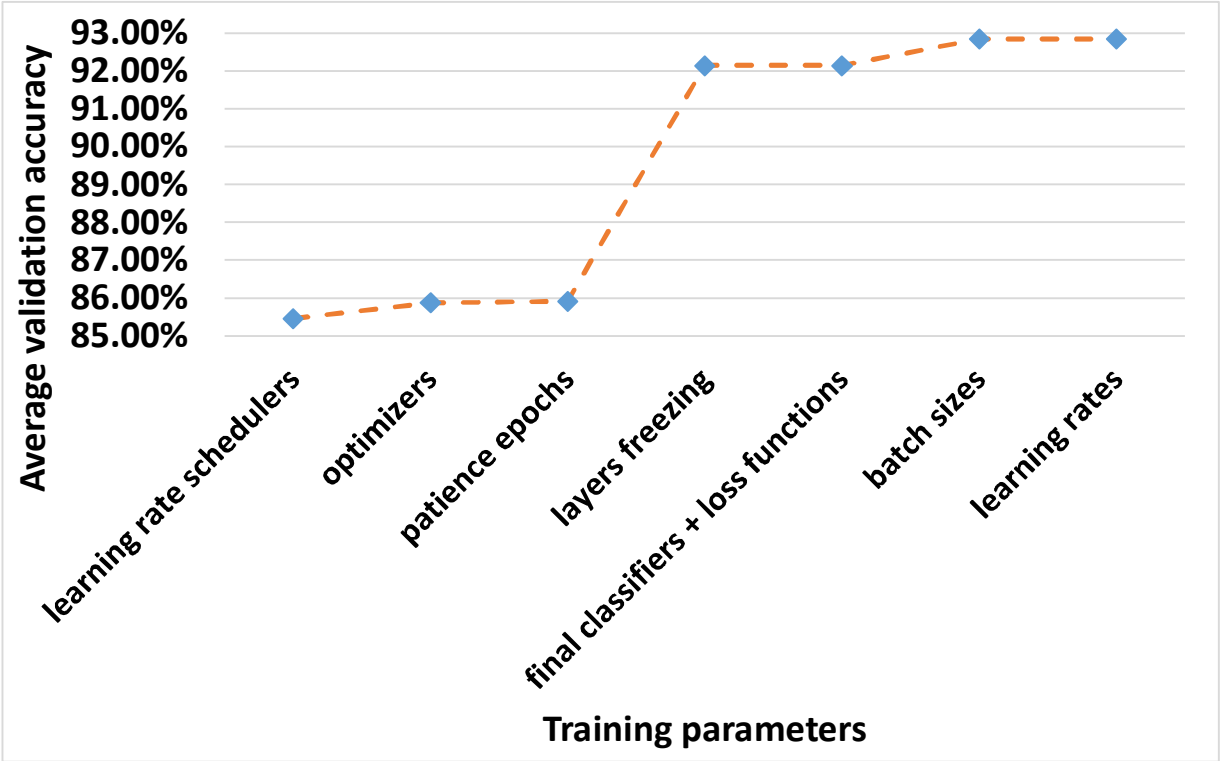

Fig. 3. The increase in validation accuracy for different training parameters

From Figure 3, the highest increase of the average validation accuracy, 7.25% increase, is conducted by tuning layers freezing parameter. The second highest increase, 0.76% increase, is achieved by tuning the batch size. In comparison, there is a big gap between the improvements achieved by the first and second most effective training parameters. On the other hand, tuning of final classifiers, loss functions, and learning rates do not have any increase on the curve.

The core functions of transfer learning, using pre-trained weights to apply knowledge of another task to the target task, can explain this significant increase conducted by tuning the layers freezing. In this case, most layers can be retrained to improve classification accuracy, which suggests a great similarity can be found between the source domain (ImageNet dataset) and the target domain (TrashNet dataset). On the other hand, If the difference between these two domains is too large, then unfreezing all layers can conduct a negative impact on the model performance since low-level features extracted are destroyed.

Next, the benefits of tuning the final classifiers, loss function, and learning rates are not shown up on the curve. This is because the Softmax classifier and Cross-Entropy loss are used as the default selection, and the other classifiers or losses do not conduct better performance. On the other hand, learning rates are selected using the learning rate finder throughout experiments. As a result, no other learning rates can be compared with to show up the benefits of tuning learning rates.

The effects of layers freezing can also be witnessed in reviewed waste classification projects. Among seven reviewed waste classification projects, six of them apply transfer learning techniques. Moreover, four of them initialise the model using pre-trained weights, and their average classification accuracy is 95.96%. However, these projects do not provide information about layers freezing, such as the number of frozen layers and position of frozen layers in the model. In this project, experiments of layers freezing are conducted, results are analysed and discussed, and the final decision on layers freezing is presented.

*2) Layers freezing*

The GoogleNet model, used in the model of paper [16], is kept as the source model of transfer learning. The only change made to the GoogleNet model structure is the final dense layer. To be specific, the number of nodes in the final dense layer is reduced from 1000 to 5 since this work aims to classify five types of recycling waste images. Next, the layers freezing experiment is carried out by freezing all layers of the transferred GoogleNet model at first. Layers are gradually unfrozen from the output dense layer to the input convolutional layer, and the one yielding higher classification accuracy are selected.

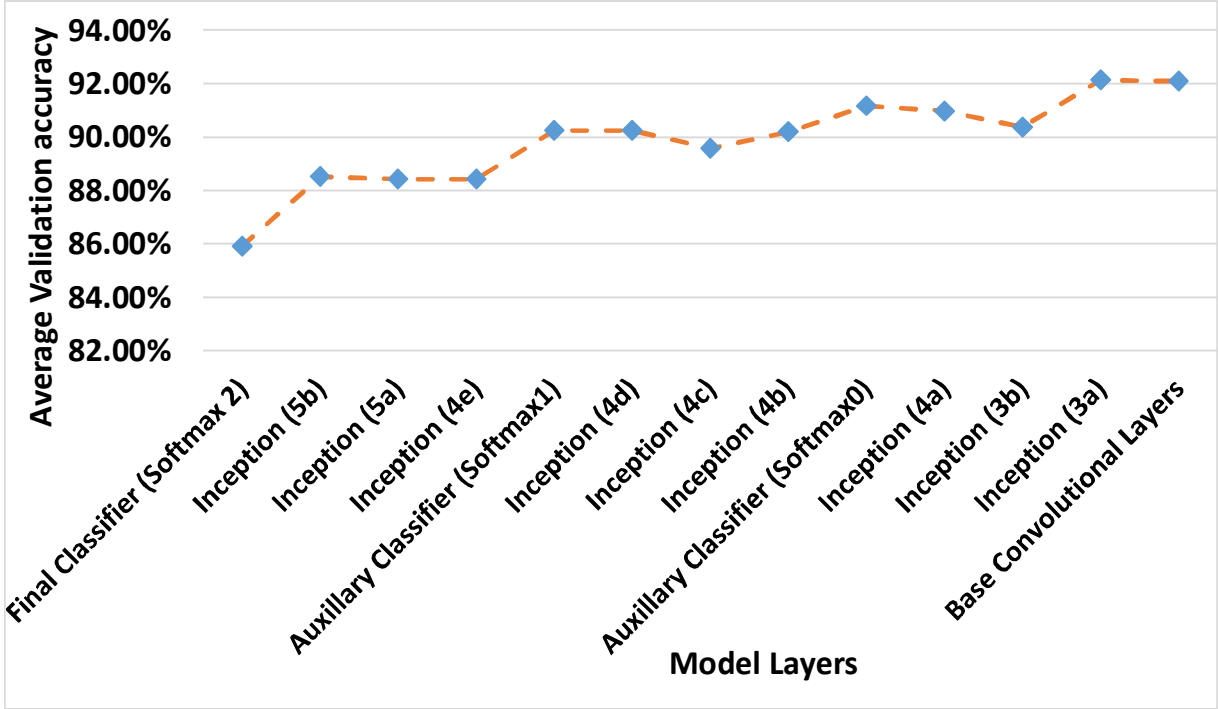

Fig. 4. Validation accuracy of different layers freezing state

From Figure 4, an increasing trend in average validation accuracy can be observed as the layers are gradually unfrozen. There are five steep increases, which happen at the unfreezing of Inception module (5b, 4b, 3a) and Auxiliary classifier (Softmax1, Softmax0). The weights of these layers are retrained from the ImageNet dataset trained weights to recognize different types of recyclables.

Among these improvements, the unfreezing of Inception module 5b increases the accuracy the most, 2.61% in accuracy. The pre-trained weights of Inception module 5b store the high-level feature knowledge, such as the general shape of a car, which are the most irrelevant knowledge of recognizing recyclables. Retraining Inception module 5b can help the model discard this unnecessary feature knowledge and learn useful feature knowledge, such as the general shape of glass bottle.

The second most increase happens at the unfreezing of Auxiliary classifier softmax1, 1.82% in accuracy. Auxiliary classifiers are used in the GoogleNet model to combat the vanishing gradient problem by calculating the final classifier loss together with auxiliary classifier losses. Retraining the auxiliary classifier softmax1 can reduce the auxiliary classifier loss and the total loss.

Although Inception module 3a is very close to the input convolutional layers, the validation accuracy still increases as the layer is unfrozen. A possible explanation is that the low-level features of the ImageNet dataset are very similar to the TrashNet dataset. On the other hand, the small accuracy reduction of base convolutional layers can be explained by the disruption of low-level feature extraction algorithm.

Overall, the layers freezing experiment helps protect the low-level feature extraction algorithm and tune the high-level feature extraction algorithm towards the recycling waste classification task. In comparison, the other waste image classification projects do not provide information about layers freezing or related experimental results. For future waste image classification projects, similar layers freezing experiments are necessary when applying transfer learning techniques.

## *C. Data augmentation*

After developing the benchmark model, various data augmentation techniques can be tested and determine which set of techniques to use. These data augmentation techniques are flipping, rotation, shear, zoom, and brightness control. In this section, data augmentation techniques used in this project are selected based on validation accuracy, and confusion matrices of different data augmentation techniques are plotted and analysed.

*1) Validation accuracy of data augmentation techniques*

The benchmark model applied with different data augmentation techniques of different magnitude. Then, the best-performed magnitude of each data augmentation technique is compared to decide the final model applied techniques.

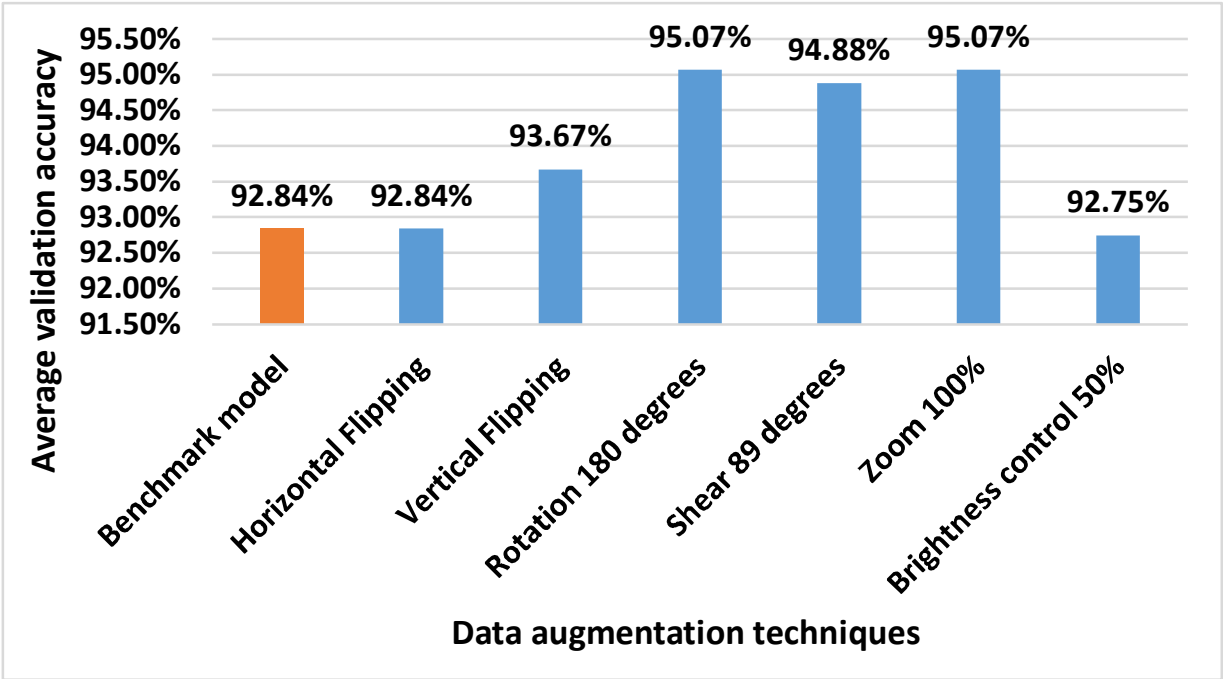

Fig. 5. Validation accuracy of different data augmentation techniques applied to the benchmark model

From Figure 5, the average validation accuracy of different data augmentation techniques applied to the benchmark model are compared and plotted in a bar chart. Besides the brightness control and horizontal flipping model, accuracy increases can be witnessed in the other models compared with the benchmark model. Unexpectedly, applying the brightness control technique conducts a small percentage decrease in validation accuracy, and this may be explained by the brightness uniformity of the TrashNet dataset. Furthermore, the TrashNet dataset is already a horizontally flipped dataset, thus, zero increase in validation accuracy in the horizontal flipping model is observed. Rotation of 180 degrees, shear of 89 degrees, and zoom of 100% have significant effects in helping the model mitigate the effects of the overfitting problem since these types of images are scarce in the TrashNet dataset.

### 2) *Confusion matrices of data augmentation techniques*

Confusion matrix of the benchmark model, final model, and the benchmark model with different data augmentation techniques applied (including flipping, zoom, rotation, and shear) are plotted. These matrices visually present the classification results of different classes, where correct predictions are plotted on the diagonal line, and wrong predictions are plotted on the remaining blocks of these matrices.

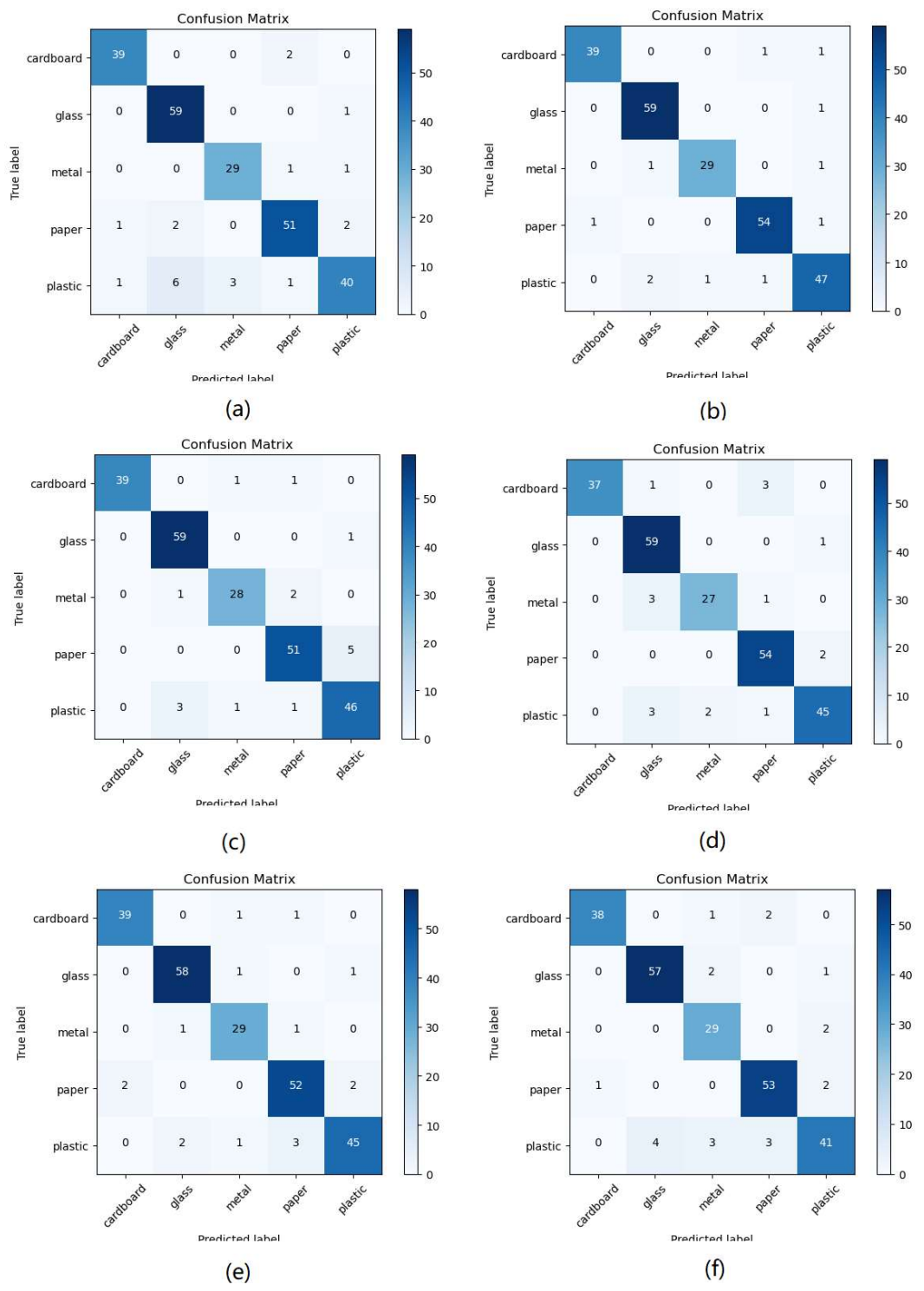

Fig. 6. Confusion matrices of different data augmentation techniques applied to the benchmark model, (a):benchmark model, (b):final model, (c):flipping, (d):zooming, (e): rotation, (f): shear

From Figure 6a, the benchmark model has the highest error rate, 21.57%, in plastic image predictions, and the second-highest error rate, 8.93%, in paper image predictions. Plastic images are often recognized incorrectly as glass or metal, and this may because they are all reflective materials. Also, paper images are often recognized incorrectly as glass and plastic, and this may because glass and plastic bottles are often surrounded by printed labels that looks like paper. Next, cardboard images are often recognized incorrectly as paper, and this may because cardboard is a paper-based product but with brownish in colour.

Figure 6b shows the final model's performance, which is applying selected data augmentation techniques to the benchmark model. To be specific, the error rate of the plastic class is reduced from 21.57% to 7.84%, and the paper class's error rate is reduced from 8.93% to 3.57%. Also, the classification accuracy of other classes remains the same. From the result, the weaknesses (recognizing paper and plastic) of this classification model are improved significantly by applying the selected set of data augmentation techniques.

Among these data augmentation techniques applied to the benchmark model, the flipping technique (shown in Figure 6c) improved the plastic class error rate by the most, from 21.67% to 9.8%. Next, the rotation and zoom techniques improve the plastic error rate slight lower, from 21.67% to 11.76%. On the contrary, the shear technique only improves the plastic error rate from 21.67% to 19.6%. This may because the flipping, rotation, and zoom technique can provide the model with additional plastic image information, such as viewing this image from another angle or focusing on smaller pixels. In comparison, the shear technique concentrates this image information by stretching this image to a parallelogram from a rectangle.

## VI. Conclusions and Future Work

This work is aimed at improving the image processing aspect of the IBM Wastenet project. Furthermore, this work provides full details of the model's training using a systematic approach, unlike other work in the field. These training procedures can be summarised as a framework for training recycling waste image classification models, including how to split the dataset, how to select training parameters, and how to choose data augmentation techniques.

The dataset is split by choosing three parameters, including split ratios, sampling strategies, and CV techniques, which are determined through statistical tests of test set mean accuracy. In this work, 10% of the entire dataset is split using the Hold-out CV technique with simple random sampling strategy, and the remaining dataset is then split into training and validation sets using the 10-fold CV technique with simple random sampling strategy.

Next, the training parameters are selected by comparing the average 10-fold validation accuracy, and these parameters are learning rate schedulers, patience epochs, layers freezing, loss function, final classifiers, batch sizes, and learning rates. In this work, the constant learning rate scheduler of 2e-5 learning rate is used together with Adam optimizer to optimise the Cross-Entropy loss. Next, the node number in the final classification layer, Softmax, of the GoogleNet model is reduced from 1000 to 5, and only base convolutional layers are frozen. Then, the pre-trained weights of the GoogleNet model are used to initialise the transferred GoogleNet model, and images are passed in a batch size of 16. Lastly, the training process is stopped if no higher validation accuracy is obtained in the next 100 epochs. The benchmark model is trained using the above-described training parameters and obtained 91.21% test set classification accuracy.

A range of data augmentation techniques are tested using the average 10-fold validation accuracy, and these augmentation techniques include flipping, rotation, shear, zoom, and brightness control. In this work, all techniques except brightness control are applied to the benchmark model, and the test set classification accuracy is increased to 95.40%. Lastly, this model has been used to perform successful real-time classification on a computer using a standard webcam.

Future work is looking at moving the developed machine learning models to the edge to overcome connectivity bottlenecks and bandwidth limitations associated with the Internet backbone. This will be achieved by performing real-time inference of a CNN on a resource-constrained device, such as NVIDIA's Jetson Nano module. Novel deep compression techniques and architectures, such as EfficientNets are also being considered to achieve accurate and fast real-time classification on such embedded devices.

## ACKNOLWEDGEMENTS

The authors are grateful to Mindy Yang and Gary Thung, who collected the TrashNet dataset and made it publicly available. The authors would also like to acknowledge the IBM UK Wastenet project team, in particular David Locke and Kaloyan Nikolaev Yordanov for their fruitful discussions.